\documentclass[english]{article}
\usepackage[latin9]{inputenc}
\usepackage[a4paper]{geometry}
\usepackage{babel}
\usepackage{textcomp}
\usepackage{amsmath}
\usepackage{amssymb}
\usepackage{graphicx}
\usepackage{microtype}
\usepackage[unicode=true,pdfusetitle,
 bookmarks=true,bookmarksnumbered=false,bookmarksopen=false,
 breaklinks=false,pdfborder={0 0 1},backref=false,colorlinks=false]
 {hyperref}

\makeatletter

\providecommand{\tabularnewline}{\\}

\usepackage{cite}
\usepackage{amsfonts}\usepackage{algorithmic}
\usepackage{textcomp}
\usepackage{xcolor}
\usepackage{microtype}

\date{}

\@ifundefined{showcaptionsetup}{}{%
 \PassOptionsToPackage{caption=false}{subfig}}
\usepackage{subfig}
\makeatother

\begin{document}
\global\long\def\model{\mathtt{ICDE}}

\title{Hybrid Generative-Discriminative Models for Inverse Materials Design}

\author{Phuoc Nguyen, Truyen Tran, Sunil Gupta, Santu Rana and Svetha Venkatesh\\
Applied AI Institute\\
Deakin University, Australia.\\
\emph{phuoc.nguyen@deakin.edu.au}}
\maketitle
\begin{abstract}
Discovering new physical products and processes often demands enormous
experimentation and expensive simulation. To design a new product
with certain target characteristics, an extensive search is performed
in the design space by trying out a large number of design combinations
before reaching to the target characteristics. However, forward searching
for the target design becomes prohibitive when the target is itself
moving or only partially understood. To address this bottleneck, we
propose to use backward prediction by leveraging the rich data generated
during earlier exploration and construct a machine learning framework
to predict the design parameters for any target in a single step.
This poses two technical challenges: the first caused due to one-to-many
mapping when learning the inverse problem and the second caused due
to an user specifying the target specifications only partially. To
overcome the challenges, we formulate this problem as conditional
density estimation under high-dimensional setting with incomplete
input and multimodal output. We solve the problem through a deep hybrid
generative-discriminative model, which is trained end-to-end to predict
the optimum design. 

\end{abstract}

\global\long\def\hb{\boldsymbol{h}}
\global\long\def\tb{\boldsymbol{t}}
\global\long\def\rb{\boldsymbol{r}}
\global\long\def\wb{\boldsymbol{w}}
\global\long\def\mb{\boldsymbol{m}}
\global\long\def\eb{\boldsymbol{e}}
\global\long\def\qb{\boldsymbol{q}}
\global\long\def\ob{\boldsymbol{o}}
\global\long\def\ab{\boldsymbol{a}}
\global\long\def\ub{\boldsymbol{u}}
\global\long\def\vb{\boldsymbol{v}}
\global\long\def\xb{\boldsymbol{x}}
\global\long\def\bb{\boldsymbol{b}}
\global\long\def\yb{\boldsymbol{y}}
\global\long\def\zb{\boldsymbol{z}}
\global\long\def\mub{\boldsymbol{\mu}}
\global\long\def\sigmab{\boldsymbol{\sigma}}
\global\long\def\cb{\boldsymbol{c}}

\section{Introduction}

\begin{figure*}
\centering{}\subfloat[Search process for simulation-based design. It may run for hours or
days per one design variable set. \label{fig:search-process}]{\begin{centering}
\includegraphics[viewport=0bp 0bp 460bp 270bp,clip,width=0.7\textwidth]{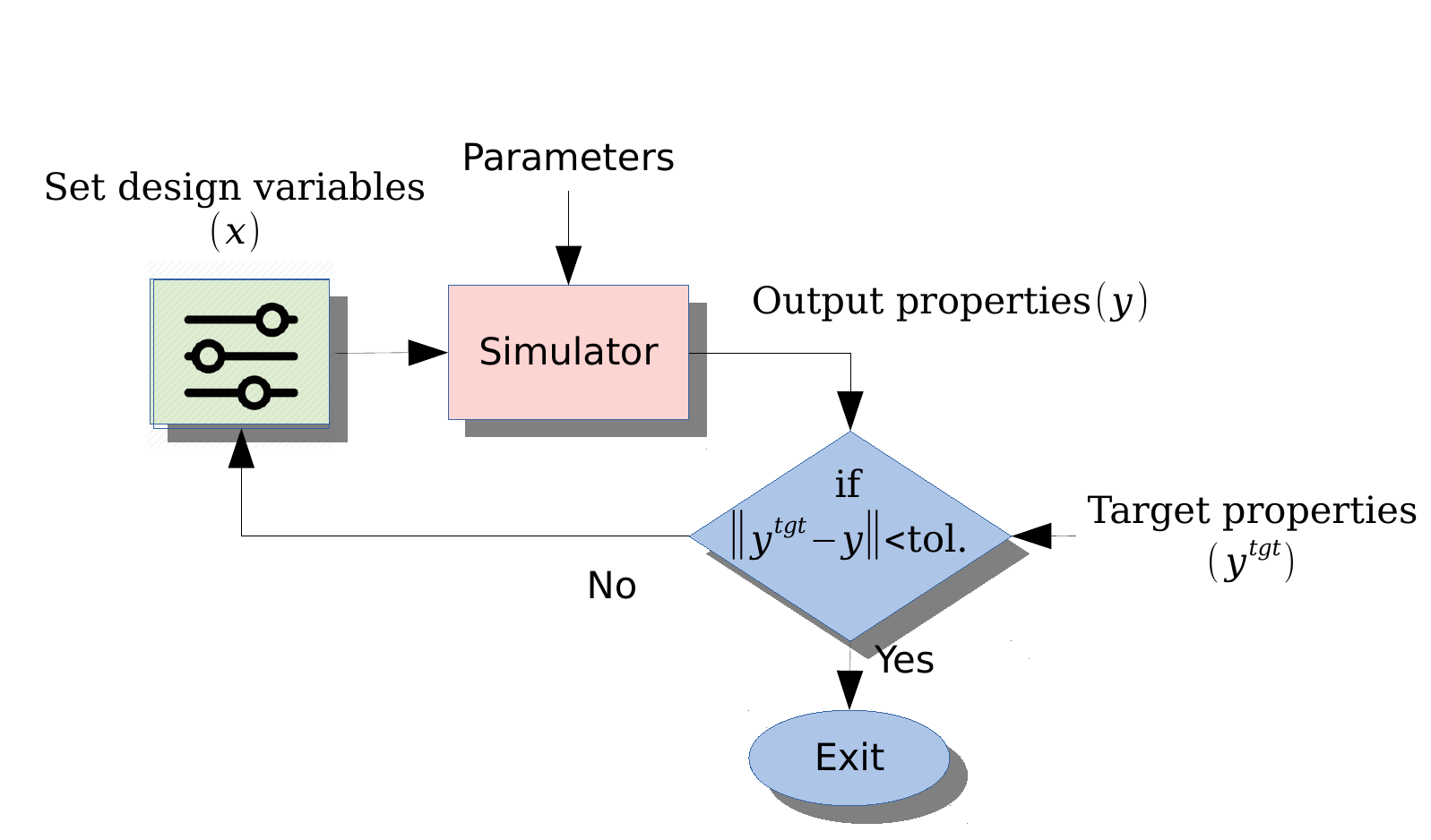}
\par\end{centering}
}\\
\subfloat[Machine learning approach, typically costs mini-seconds per tens of
predictions. \label{fig:ml-process}]{\begin{centering}
\includegraphics[viewport=0bp 0bp 460bp 270bp,clip,width=0.7\textwidth]{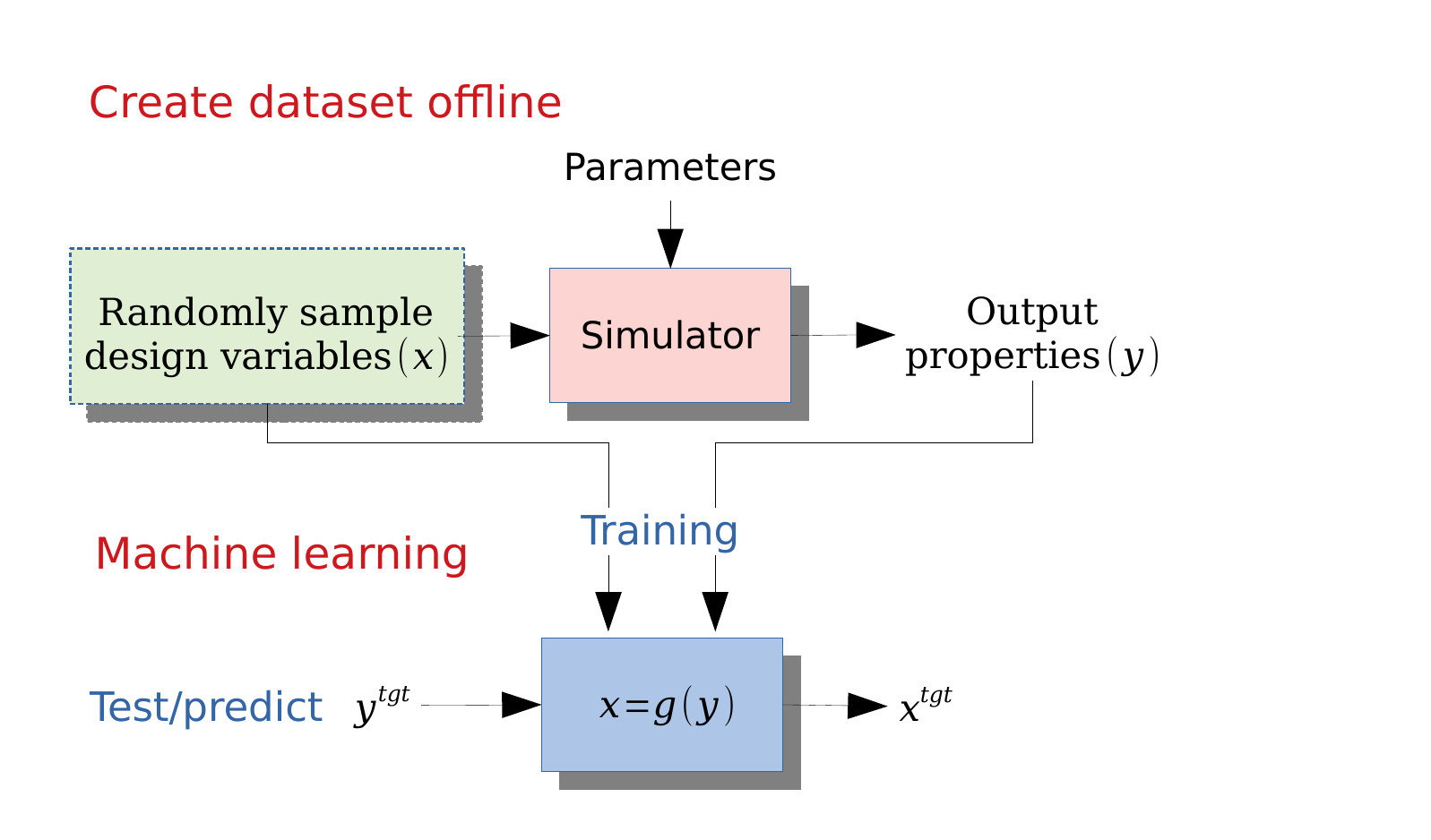}
\par\end{centering}
}\caption{Experimental design paradigms for design and discovery of new products.
(a) current design paradigm requiring a long iterative search procedure
(b) The proposed design paradigm. }
\end{figure*}

Scientific innovations relating physical processes require laborious
experimentation and expensive simulation as the relationships between
design variables and output characteristics are unknown \cite{gani2002property,sanchez2018inverse}.
To design a new product with certain target characteristics, a search
is typically performed in the design space -- a large number of the
design combinations (input variables) are tried in simulators before
reaching to the target characteristics (see Fig.~\ref{fig:search-process}).
Although modern search methods such as Bayesian Optimization \cite{shahriari2016taking}
are efficient, there is an inherent problem with the search: Each
time the target characteristics are changed or refined, the search
process needs to be restarted making the design task extremely time-consuming.
The current search paradigm does not harness existing experimentation
data and simulator queries systematically as there is no clear provision
to reuse them.

Thanks to the availability of growing data, accurate simulators and
modern machine learning algorithms, this search process can be avoided
with a potential to accelerate scientific innovations by multiple
orders of magnitude. An effective way to address this problem is to
leverage the existing data and query the simulators in an offline
mode to sample the data space sufficiently and then harness this data
to build a machine learning model. Given sufficient data, modern machine
learning (ML) models (e.g. a deep neural network) can approximate
the underlying physical relationships and the simulators arbitrarily
closely. The ML models can then be used to convert the search process
in a less expensive optimization. Denoting the function learned by
ML model as $f(\boldsymbol{x})$, we can discover the target design
by solving the following optimization problem: $\boldsymbol{x}^{\text{target}}=\underset{\boldsymbol{x}\in\mathcal{X}}{\text{argmin}}\left\Vert f(\boldsymbol{x})-\boldsymbol{y}^{\text{target}}\right\Vert $. 

Although the above--mentioned approach can replace the expensive
search via experimentation and simulators, we still need to solve
a global optimization problem each time we have a specified design
goal $\boldsymbol{y}^{\text{target}}$. Fortunately, we can also avoid
the global optimization by simply flipping our problem of learning
$f(\boldsymbol{x})$ to the learning of its \emph{inverse design function}
$g(\boldsymbol{x})=f^{-1}(\boldsymbol{x})$ \cite{sanchez2018inverse}.
This offers a paradigm shift in modeling as by learning the function
$\boldsymbol{x}=g(\boldsymbol{y})$, we can directly predict our design
variables in a single step as $\boldsymbol{x}^{\text{target}}=g(\boldsymbol{y}^{\text{target}})$
without needing any search or global optimization (see Fig.~\ref{fig:ml-process}). 

However when taking this approach to designing new products, we face
two technical difficulties. The first difficulty is that the forward
function $f$ may be many-to-one in some problem domains, \emph{i.e.}
it might be possible to have same output for many input combinations.
In such cases, the inverse function will be ill-posed as there would
be multiple $\boldsymbol{x}$ solutions for the same $\boldsymbol{y}$.
Another practical difficulty that arises when designing new product
and processes is partial specification of the target characteristics.
This happens mainly for two reasons: the first, an user may not be
absolutely certain about setting all the target characteristics, instead
he/she might want to leave some of the target characteristics free;
and the second, he/she may not know the precise value of a certain
target characteristic. Due to these $\boldsymbol{y}^{\text{target}}$
may only be partially specified for a new product.

We overcome the difficulties by formulating this inverse design problem
as an instance of conditional density estimation under high-dimensional
setting with incomplete input and multimodal output. We solve the
problem by designing a deep hybrid generative-discriminative model,
which integrates Mixture Density Networks (MDN) \cite{bishop1994mixture}
and Conditional Deep Generative Models (CVAE \cite{sohn2015learning}
and CGAN \cite{mirza2014conditional}). The MDN component addresses
the first difficulty allowing multiple designs for a target. The CVAE/CGAN
component addresses the second difficulty by imputing the missing
target specifications through the data distribution. Putting together
in the integrated model permits multiple design possibilities catering
for the degree of freedom caused by either the many-to-one forward
function and the incomplete specification. As a result, we gain the
freedom to discover a whole sub-class of products meeting the partial
target specifications.

We focus on an alloy design problem and use a metallurgical model
known as CALPHAD implemented in the Thermo-Calc software \cite{kaufman1970computer,lukas2007computational}.
The input to our model is a partial phase diagram and the output is
an alloy composition. Our goal is to specify a desired phase diagram
that is associated to different alloy properties, e.g., its strength,
weld-ability, its corrosion resistance etc., and predict an alloy
elemental composition that is highly likely to form the specified
phases and therefore show the intended properties. We carried out
experiments on two datasets acquired by querying the Aluminum alloy
database. The first dataset was created from 30 Aluminum alloys\footnote{Alloy IDs: 2014, 2018, 2024, 2025, 2218, 2219, 2618, 6053, 6061, 6063,
6066, 6070, 6082, 6101, 6151, 6201, 6351, 6463, 6951, 7001, 7005,
7020, 7034, 7039, 7068, 7075, 7076, 7175, 7178, 7475.}. The second dataset was created by using Bayesian Optimization and
searching for alloys satisfying a common FCC (Face-Centered Cubic)\footnote{FCC is a ``crystal structure consisting of an atom at each cube corner
and an atom in the center of each cube face'' {[}corrosionpedia.com{]}.} property existed in the 30 mentioned Aluminum alloys. We vary the
composition of each alloy in both sets in a window of $\pm20\%$ to
derive necessary data to train and test our models. We show that our
model can accurately predict the alloy compositions with an average
relative error of $1.8\%$ and $2.95\%$ for the first and second
datasets respectively. The main contributions of this paper are: 
\begin{itemize}
\item A new, efficient data-driven approach to design targeted products.
This is done by turning the traditional simulation-based search for
the required target into a direct prediction using a multimodal inverse
function.
\item A deep hybrid generative-discriminative model to solve the resulting
problem of conditional density estimation with incomplete input and
multimodal output. The model is capable of taking partial target specification
to predict a class of product designs. 
\item Demonstration of our approach for alloy design in an entirely novel
way with a possible speed up by hundreds of thousands of times\footnote{Code and data download link: https://bit.ly/2C2HgdV}.
\end{itemize}
Our work can be seen as one of the early data mining work in the emerging
field of materials informatics \cite{agrawal2016perspective,kalidindi2015materials,liu2017materials},
the need of which has been greatly emphasized by the Materials Genome
Initiative (MGI) project\footnote{https://www.mgi.gov}. The significance
of our work lies in addressing the limitation of the third paradigm
of materials discovery (using simulation) through through the use
of data--intensive methodologies, which are in line with the current
shift into the fourth paradigm of science discovery \cite{hey2009fourth}.

\section{Related Work}

\paragraph{Data-driven materials discovery}

The third paradigm in science discovery involves computer simulation
of physical processes \cite{hey2009fourth}. Simulators offer a high-throughput
test-bed to perform a search or optimization to discover products
with intended target properties \cite{li2012using,norskov2011density,ramprasad2017machine,weissman2014design}.
Collective effort in these experimental and simulation activities
in materials science has led to availability of large repositories
\cite{ramprasad2017machine}, e.g., NOMAD\footnote{http://nomad-coe.eu},
Materials Project\footnote{http://materialsproject.org}, Aflowlib\footnote{http://www.aflowlib.org},
and OQMD\footnote{http://oqmd.org}. These databases specify the structure-property
relationships in materials. This has given rise to the fourth paradigm
of data--intensive discovery using machine learning \cite{agrawal2016perspective,hey2009fourth,kalidindi2015materials,liu2017materials,pham2018graph,raccuglia2016machine}
to learn to predict the structure-property relationships from data
without expensive simulation. The prediction is orders of magnitude
faster than the traditional simulation and experimental work required
to generate the data in the first place. The prediction can be used
in an inner loop of an optimization process to search for the best
design.

\paragraph{Inverse materials design}

The traditional materials discovery loop consists of several steps
\cite{sanchez2018inverse}: (i) concept generation, and simulation-based
screening; (ii) material synthesis; (iii) measuring wanted properties
in the context of use. The cycle is repeated and materials are refined
until design criteria are met. It takes many years or even decades
in the process. \emph{Inverse design} offers an alternative route
by inverting this conventional paradigm, starting from the target
functionalities and search for a perfect design \cite{zunger2018inverse,aspuru2018materials}.
Recent works that employ deep generative models (such as variational
autoencoder \cite{kingma2013auto} and generative adversarial networks
\cite{goodfellow2014generative}), coupled with Bayesian optimization
\cite{shahriari2016taking}, have proved to be successful in generating
molecules and materials \cite{gomez2018automatic}.

However, on crucial drawback still remains. Even with machine learning-based
surrogate models to replace simulation, this optimization loop is
still expensive especially when the target functionalities are constantly
changing. To the best of our knowledge, our work is the first effort
to completely eliminate the search/optimization involved in materials
discovery process. Our inverse modeling can predict the intended design
configuration instantly in a single step.

\paragraph{Imprecise specification}

The problem of imprecise output specification has been discussed before
in a limited context. For example, \cite{qian1993process,limbourg2005optimization}
assume a misspecification of output due to a small uncertainty. However,
none of the prior work allow the output specifications to be partially
expressed or use imprecise specification to explore a class of products.
Our approach to handle partial output specifications uses conditional
deep generative models \cite{sohn2015learning,mirza2014conditional},
similar to recent work in data imputation \cite{mattei2018leveraging,nazabal2018handling,rezende2014stochastic,yoon2018gain}.
The main architectural difference is our integration of imputation
to discriminative models. The discriminative component in our model
is is related to recent deep nets for conditional density estimation
(e.g., see \cite{le2018variational,shu2017bottleneck,trippe2018conditional}
for recent works and references therein). The main difference is we
have incomplete input and multimodal output.

\section{Preliminaries \label{sec:Prelim}}

\subsection{Alloys phase diagrams}

Alloys represent a huge class of metals being used in all walks of
modern life. An alloy composition consists of a main metal with highest
proportion and the remaining metal elements accounting for the rest.
We primarily work on Aluminum alloy here, and the other elements used
in this research are Cr, Cu, Mg, Ti, Zn, Zr, Mn, Si, and Ni. 

\begin{figure}
\begin{centering}
\begin{tabular}{c}
\includegraphics[width=1\columnwidth]{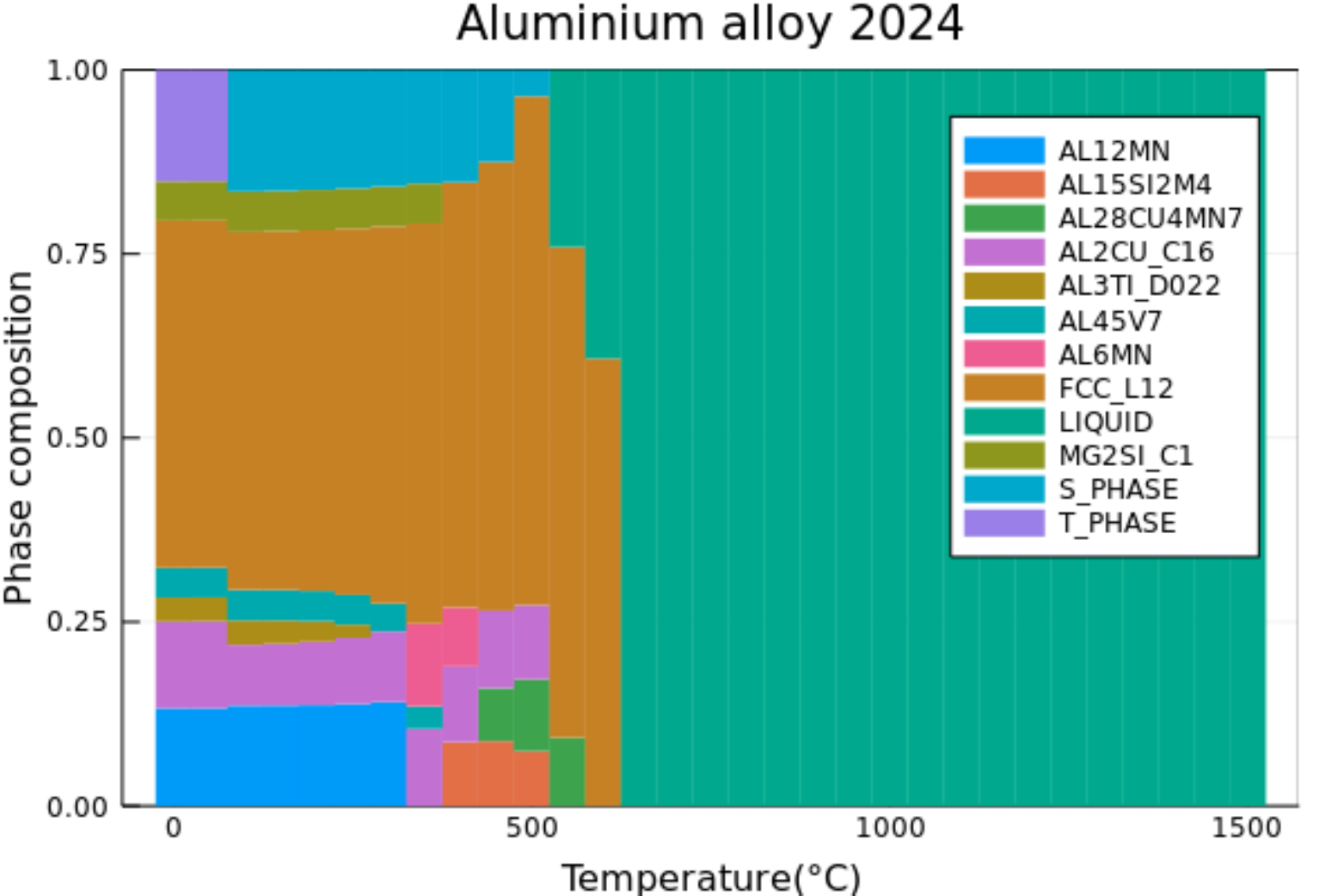}\tabularnewline
\end{tabular}
\par\end{centering}
\caption{Phase diagram alloy 2024. The phases are square-root scaled then re-normalized
for plotting purpose. Best viewed in color. \label{fig:phase-diagrams}}
\end{figure}

A phase is a state of matter. The states of a material, especially
alloys, at a given temperature and pressure may include multiple phases.
The collection of phase distributions across temperatures and pressures
constitute a phase diagram. Phase diagram can be used as a proxy for
materials properties because materials designers can infer target
properties from it. Phase diagrams can be computed using software
like Thermo-Calc\footnote{http://www.Thermo-Calc.com}. We give as
input the elements and its fractions into the Thermo-Calc software
and get it to simulate and generate distribution of the metallic phases
across temperatures as output. Fig.~\ref{fig:phase-diagrams} shows
a colored form of a phase diagram of Aluminum alloy No. 2024 under
atmosphere pressure. As a result, they have a mixture of metallic
phases up to the melting point for all compounds. For example Aluminum
alloy 2024 has 7 metallic phases at temperature range 0-300 °C, then
6, 5, 3, and 2 metallic phases at temperature 350, 400-500, 550, and
600 °C respectively, and finally only 1 liquid phase from 650 °C onward.
The FCC (which stands for Face-Centered Cubic) phase is the largest
solid phase, which will be of primary interest in our empirical study
in Section~\ref{sec:Experiments}. 

\subsection{Inverse problems \label{subsec:Inverse-problems}}

Let $\xb\in\mathcal{X}$ be a vector describing the design parameters,
and $\yb$ be a set of target properties, which, in the case of materials
design, can have mechanical, thermal, optical, chemical or biological
characteristics. As an example, in alloy discovery, the design parameters
can be the mixture of components of the alloy, and the target properties
can be melting temperature, hardness, elasticity and surface resistance
against corrosion. Physical laws dictate that there exist a function
of the form $\yb=f(\xb)$, which is known as structure-property linkage
\cite{gupta2015structure}. However, simulation of realistic matters
is extremely complex\footnote{Quantum computation using DFT takes hours to compute chemical properties
of small molecules, but it is practically impossible to compute for
just one micro-cube of matters even using the best supercomputer.}, and thus accurate simulation of $f(\xb)$ often demands great computational
and time resources \emph{for each} $\xb$. This may prevent comprehensive
exploration of the huge design space to reach a desirable target. 

Fortunately, collective worldwide effort has generated large databases
of structure-property pairs for known materials. Machine learning
offers an alternative data--driven approach to vastly speed up this
exploration \cite{gupta2015structure}. On one hand, we can approximate
the function $f(\xb)$ by learning a variational fast alternative
$f_{\pi}(\xb)$ parameterized by $\pi$ (e.g., weights of a deep neural
net). Learning occurs only once on a training set $D=\left\{ \left(\xb_{i},\yb_{i}\right)\right\} $,
where $\xb_{i}$ is a sample in $\mathcal{X}$ and $\yb_{i}$ is computed
by running the simulator. Once the function has been estimated, the
search of a desirable target $\yb^{target}$ can be efficiently carried
out through global optimization $\xb^{target}=\text{argmin}_{\xb\in\mathcal{X}}\left\Vert f_{\pi}(\boldsymbol{x})-\boldsymbol{y}^{\text{target}}\right\Vert $.

\section{Methods \label{sec:Methods}}

In this section, we present our contributions in solving inverse-problems
in science discovery through a hybrid generative-discriminative model
that accounts for conditional density estimation under incomplete
input and multimodal output. 

\subsection{Inverse-problem as data--driven inference\label{subsec:Inverse-problem-as-data=002013driven}}

We propose eliminate the global optimization stated in Sec.~\ref{subsec:Inverse-problems}
entirely by estimating an \emph{inverse-function} of $f$ using $g_{\eta}(\yb)$
such as that $g_{\eta}(\yb_{i})\approx\xb_{i}$ for all $i$. With
this inverse-function, searching for target designs $\xb^{target}$
to meet target properties $\yb^{target}$ is instant. However, estimating
$g_{\eta}(\yb)$ is challenging for the following reasons. First,
the inverse function $f^{-1}$ may not exist since $f$ can be a many-to-one
mapping, that is, one target output $\yb$ can be satisfied by multiple
input designs $\xb$. Second, in practice the target output may not
be fully specified by the scientist. Sometimes this incomplete specification
can be due to ignorant (we do not know exactly what we want in the
first place), or on purpose (we want not an exact design, but a class
of designs). A partial specification of $\yb$ further increases the
uncertainty about the input design $\xb$. Let $\yb=\left(\vb,\hb\right)$
where $\vb$ denotes the specified target component and $\hb$ the
unspecified counterpart. Given the uncertainty in the target designs
$\xb$, \emph{the inverse-problem is best reformulated as estimating
an entire conditional density spectrum} $P\left(\xb\mid\vb\right)$. 

\subsection{Multimodal density estimation given incomplete conditions \label{subsec:Conditional-density-estimation}}

Assume that $\hb$ is generated by a latent variable $\zb$. We aim
to model a conditional density of the design $\xb$ given the specified
$\vb$ target component:
\begin{equation}
P\left(\xb\mid\vb\right)=\int_{\hb}\int_{\zb}P\left(\xb,\hb,\zb\mid\vb\right)d\hb d\zb\label{eq:model}
\end{equation}
The joint density $P\left(\xb,\hb,\zb\mid\vb\right)$ is further factorized
as:
\begin{equation}
P\left(\xb,\hb,\zb\mid\vb\right)=P\left(\xb\mid\vb,\hb\right)P\left(\hb\mid\vb,\zb\right)P\left(\zb\mid\vb\right)\label{eq:factorize}
\end{equation}
This leads to
\begin{align}
P\left(\xb\mid\vb\right) & =\int_{\hb}\int_{\zb}P\left(\xb\mid\vb,\hb\right)P\left(\hb\mid\vb,\zb\right)P(\zb\mid\vb)d\hb d\zb\nonumber \\
 & \approx\frac{1}{N}\sum_{i=1}^{N}\mathbb{E}_{P\left(\hb\mid\vb,\zb^{(i)}\right)}\left[P\left(\xb\mid\vb,\hb\right)\right]\label{eq:predict-x}
\end{align}
where $\zb\sim P\left(\zb\mid\vb\right)$.

However, integrating over $\hb$ is still intractable. Assume further
that the imputation of the missing component via $P\left(\hb\mid\vb,\zb\right)$
takes a simple Gaussian form, that is, $\hb\sim\mathcal{N}\left(\mub(\vb,\zb);I\sigma^{2}(\vb,\zb)\right)$.
For simplicity, we approximate the expectation $\mathbb{E}_{P\left(\hb\mid\vb,\zb^{(i)}\right)}\left[f(\hb)\right]$
by a function evaluation at the mode, i.e., $\mathbb{E}_{P\left(\hb\mid\vb,\zb^{(i)}\right)}\left[f(\hb)\right]\approx f\left(\bar{\hb}\right)$
where $\bar{\hb}=\mub(\vb,\zb)$. This leads to
\begin{equation}
P\left(\xb\mid\vb\right)\approx\frac{1}{N}\sum_{i=1}^{N}P\left(\xb\mid\vb,\mub\left(\vb,\zb^{(i)}\right)\right)\label{eq:predict}
\end{equation}

\begin{figure}
\noindent \begin{centering}
\includegraphics[viewport=60bp 15bp 260bp 120bp,clip,scale=0.7]{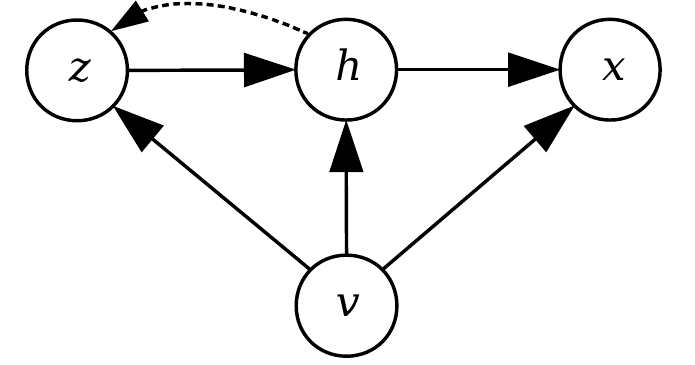}
\par\end{centering}
\caption{Graphical model of the proposed method. $\protect\xb$: input design,
$\protect\vb$: specified target component, $\protect\hb$: unspecified
part, and $\protect\zb$: latent variable.\label{fig:Graphical-model}}
\end{figure}

As $\mub\left(\vb,\zb^{(i)}\right)$ serves as a reconstruction of
the missing component $\hb$, this somewhat resembles the technique
of multiple imputations in the literature \cite{schafer1999multiple}.
The graphical model of the factorization in Eq.~(\ref{eq:factorize})
is depicted in Fig.~\ref{fig:Graphical-model}. 

\subsubsection{Multimodal density}

Due to the one-to-many mapping in the inverse problem, the conditional
density $P\left(\xb\mid\vb,\bar{\hb}\right)$ may be multimodal, that
is, we may have multiple classes of solution. For example, in the
case of alloy design, this is naturally the case as alloys seem to
concentrate around rather separate modes each of which has a dominant
metal. To account for multiple modes, we propose to use the mixture
model:

\begin{equation}
P_{\gamma}\left(\xb\mid\vb,\bar{\hb}\right)=\sum_{k=1}^{K}\alpha_{k}P_{k}\left(\xb\mid\vb,\bar{\hb}\right)\label{eq:MDN}
\end{equation}
with mixture components $\alpha_{k}$ subject to $\alpha_{k}\ge0$
and $\sum_{k=1}^{K}\alpha_{k}=1$. 

The mixture components are implemented using a softmax neural network
as follows: 
\[
\alpha_{k}=\frac{\exp\left(f_{k}\left(\vb,\bar{\hb}\right)\right)}{\sum_{j}\exp\left(f_{j}\left(\vb,\bar{\hb}\right)\right)}
\]
where $f_{k}$ are feedforward neural networks. 

Using the re-parametrization trick (e.g., see \cite{kingma2013auto})
we model $P_{k}\left(\xb\mid\yb\right)$ as a Gaussian of mean $\mub_{k}=g_{k}^{\mu}\left(\vb,\bar{\hb}\right)$
and isotropic covariance matrix $I\sigma_{k}^{2}$ for identity matrix
$I$ and $\sigma_{k}^{2}=\exp\left(g_{k}^{\sigma}\left(\vb,\bar{\hb}\right)\right)$.
Here $g_{k}^{\mu}\left(\vb,\bar{\hb}\right)$ and $g_{k}^{\sigma}\left(\vb,\bar{\hb}\right)$
are parametrized as deep neural networks, making this model resemble
the mixture density network (MDN) \cite{bishop1994mixture}.

\subsection{Hybrid generative-discriminative learning}

\subsubsection{CVAE imputation}

Now what remains is a model to account for the generation of the missing
target $P\left(\hb\mid\vb\right)=\int_{\zb}P\left(\hb\mid\vb,\zb\right)P\left(\zb\mid\vb\right)d\zb$.
One popular solution is Conditional Variational AutoEncoder (CVAE)
\cite{sohn2015learning}, which maximizes the lowerbound of $\log P\left(\hb\mid\vb\right)$:

\begin{align}
\mathcal{L}_{\text{CVAE}}\left(\theta,\phi\right) & =-D_{\text{KL}}\left(Q_{\phi}\left(\zb\mid\hb,\vb\right)\|P\left(\zb\right)\right)+\mathbb{E}_{z\sim Q_{\phi}}\left[\log P_{\theta}\left(\hb\mid\zb,\vb\right)\right]\label{eq:CVAE}
\end{align}
where $Q_{\phi}\left(\zb\mid\hb,\vb\right)$ denotes the recognition
model that approximates the posterior $P\left(\zb\mid\hb,\vb\right)$.
More precisely $Q_{\phi}\left(\zb\mid\hb,\vb\right)$ is a multivariate
Gaussian of mean $\mub_{\phi}\left(\hb,\vb\right)$ and diagonal covariance
$I\sigma_{\phi}^{2}\left(\hb,\vb\right)$, where $\mub_{\phi}\left(\hb,\vb\right)$
and $\sigma_{\phi}^{2}\left(\hb,\vb\right)$ are neural networks.
Hence sampling from $Q_{\phi}$ is straightforward, that is $\zb=\mub_{\phi}\left(\hb,\vb\right)+\sigma_{\phi}\left(\hb,\vb\right)\boldsymbol{\epsilon}$
for $\boldsymbol{\epsilon}\sim{\cal N}\left(\boldsymbol{0},I\right)$.

Finally, both the MDN in Eq.~(\ref{eq:MDN}) and CVAE in Eq.~(\ref{eq:CVAE})
can be jointly trained by maximizing the following hybrid objective:

\begin{equation}
\mathcal{L}\left(\theta,\phi,\gamma\right)=\mathcal{L}_{\text{CVAE}}+\lambda\mathbb{E}_{z\sim Q_{\phi}}\left[\log P_{\gamma,\theta}\left(\xb\mid\vb,\zb\right)\right]\label{eq:cvae+mlp/mdn}
\end{equation}
for some $\lambda>0$ to ensure the matching scale for both objective
terms. Note that the density function $P_{\gamma,\theta}\left(\xb\mid\vb,\zb\right)$
has two parameter sets $(\gamma,\theta)$, where $\theta$ is from
the network underlying the generation model $P_{\theta}\left(\hb\mid\zb,\vb\right)$
(as part of CVAE in Eq.~(\ref{eq:CVAE})) and $\gamma$ is from the
network underlying the multimodal density function $P_{\gamma}\left(\xb\mid\vb,\bar{\hb}\right)$
(as part of MDN in Eq.(\ref{eq:MDN})). The two networks connect through
the mean function $\bar{\hb}=\mub(\vb,\zb)$. This enables backprop
of gradient from $\xb$ to $\vb$.

\subsubsection{CGAN imputation}

An alternative to CVAE is Conditional Generative Adversarial Networks
(CGAN) \cite{mirza2014conditional}, where the loss is:
\begin{equation}
\mathcal{L}_{\text{CGAN}}\left(\theta,\eta\right)=\mathbb{E}_{\hb\sim P_{\text{data}}(\hb)}\left[\log D_{\eta}\left(\hb\mid\vb\right)\right]+\mathbb{E}_{\zb\sim P(\zb\mid\vb)}\left[\log\left(1-D_{\eta}\left(G_{\theta}(\zb,\vb)\mid\vb)\right)\right)\right]\label{eq:CGAN}
\end{equation}
where $G_{\theta}(\zb,\vb)$ is a deterministic function to generate
$\hat{\hb}$ given random prior sample $\zb$ and given specificiation
$\vb;$$D_{\eta}\left(\hb\mid\vb\right)\in(0,1)$ is discriminator
which returns high value if $\hb$ is drawn from the real data, and
low value otherwise. The solution is $\arg\min_{\theta}\arg\max_{\eta}\mathcal{L}_{\text{CGAN}}\left(\theta,\eta\right)$. 

Putting together with the MDN in Eq.~(\ref{eq:MDN}), we have the
following hybrid objective:
\begin{equation}
\mathcal{L}\left(\theta,\eta,\gamma\right)=-\mathcal{L}_{\text{CGAN}}+\lambda\mathbb{E}_{z\sim P(\zb)}\left[\log P_{\gamma,\theta}\left(\xb\mid\vb,\zb\right)\right]\label{eq:cgan+mlp/mdn}
\end{equation}
whose solution is $\arg\max_{\theta,\gamma}\arg\min_{\eta}\mathcal{L}\left(\theta,\eta,\gamma\right)$.
Note several differences from the CVAE-based hybrid objective in Eq.~(\ref{eq:cvae+mlp/mdn}).
Apart from the signature minmax solution in GAN, $\zb$ is drawn from
the prior $P(\zb)$ instead of the posterior $Q_{\phi}\left(\zb\mid\hb,\vb\right)$.

Although the original GAN in \cite{goodfellow2014generative} is known
to suffer from mode collapse(i.e., it generates the same $\hb$ for
any $\zb\sim P(\zb)$), we did not observe the phenomenon in our case.
Possibly it is because of the discriminative objective that prevents
$\hat{\hb}=G_{\theta}(\zb,\vb)$ from collapsing.

\subsubsection{Remarks}

We noted that there are more advanced versions based on the normalizing
flow framework \cite{rezende2015variational} where the Gaussian distribution
is replaced by a more complex posterior. Our framework can be directly
extended to use these more flexible approximation techniques but we
leave this for future work and consider here the basic variational
version. 

\subsection{Prediction}

At test time, we often wish to sample specific design parameters from
the conditional distribution $P\left(\xb\mid\vb\right)$ given in
Eq.~(\ref{eq:predict}). For each sample of the prior $\zb$ we have
a MDN of the form $P\left(\xb\mid\vb,\mu\left(\vb,\zb^{(i)}\right)\right)$.
Since each MDN is a mixture model, there are also multiple ways to
sample $\xb$ for each $\zb_{j}$. For evaluation, we use Gumbel sampling
\cite{maddison2014sampling} for identify major modes of significant
mass for each mixture model. The collection of all such modes constitutes
our prediction set.

\section{Experiments \label{sec:Experiments}}

We now demonstrate the effectiveness of the new methods CVAE+MDN presented
in Section~\ref{sec:Methods} in the domain of alloy discovery. In
particular, we focus on Aluminum alloys, i.e., those mixed materials
in which aluminum (Al) is the predominant metal. This represents one
of the most important classes of alloys widely used in engineering
and everyday products thanks to its light weight or corrosion resistance.
Table~\ref{tab:alloy-compositions} shows examples of element compositions
of 4 alloys, with IDs 2024, 2025, 6061 and 6066, respectively. 

\begin{table}
\begin{centering}
\par\end{centering}
\begin{centering}
\begin{tabular}{|c|c|c|c|c|}
\cline{2-5} 
\multicolumn{1}{c|}{} & \multicolumn{4}{c|}{\textbf{Alloy ID}}\tabularnewline
\cline{2-5} 
\multicolumn{1}{c|}{} & \textbf{2024} & \textbf{2025} & \textbf{6061} & \textbf{6066}\tabularnewline
\hline 
\textbf{Cr} & 0.05 & 0.05 & 0.2 & 0.2\tabularnewline
\hline 
\textbf{Cu} & 4.35 & 4.45 & 0.275 & 0.95\tabularnewline
\hline 
\textbf{Mg} & 1.5 & 0 & 1 & 1.1\tabularnewline
\hline 
\textbf{Ti} & 0.05 & 0.05 & 0.05 & 0.1\tabularnewline
\hline 
\textbf{Zn} & 0.1 & 0.1 & 0.1 & 0.1\tabularnewline
\hline 
\textbf{Zr} & 0 & 0 & 0 & 0\tabularnewline
\hline 
\textbf{Mn} & 0.6 & 0.8 & 0.05 & 0.85\tabularnewline
\hline 
\textbf{Si} & 0.1 & 0.85 & 0.6 & 1.35\tabularnewline
\hline 
\textbf{Ni} & 0 & 0 & 0 & 0\tabularnewline
\hline 
\textbf{Al} & 93.25 & 93.7 & 97.725 & 95.35\tabularnewline
\hline 
\end{tabular}
\par\end{centering}
\caption{Composition proportions (\%) of four example alloys. \label{tab:alloy-compositions}}
\end{table}

\subsection{Datasets}

\begin{figure}
\begin{centering}
\includegraphics[width=1\columnwidth]{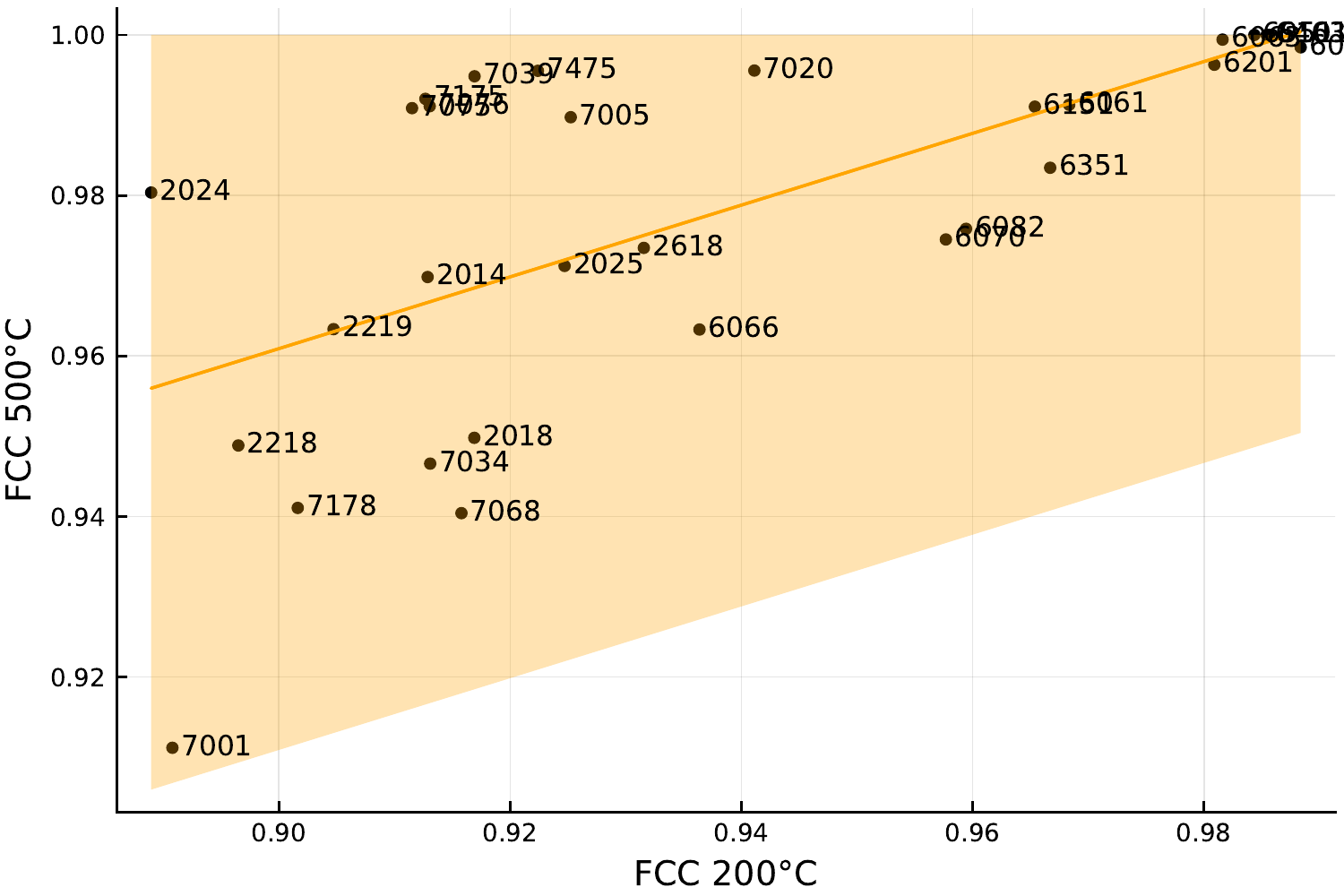}
\par\end{centering}
\caption{FCC phase at 200°C and 500°C of 30 known alloys. \label{fig:fcc-phases}}
\end{figure}

We created two datasets for our experiments
\begin{itemize}
\item \textbf{Neighborhood of known alloys}: The first dataset is of size
15,000 alloys from the following 30 known series of Aluminum alloys:
2014, 2018, 2024, 2025, 2218, 2219, 2618, 6053, 6061, 6063, 6066,
6070, 6082, 6101, 6151, 6201, 6351, 6463, 6951, 7001, 7005, 7020,
7034, 7039, 7068, 7075, 7076, 7175, 7178, and 7475. The dataset was
generated as follows. First, for each base alloy, we varied its 9
auxiliary elements (Cr, Cu, Mg, Ti, Zn, Zr, Mn, Si, Ni) by a relative
amount of $\pm20\%$ and make sure that the total proportion of the
10 elements is 100\%. Second, we fed these alloy compositions into
the Thermo-Calc software and ran the physical simulation. The simulation
pressure parameter was set to 1 atmosphere, and temperature was varied
from 0 to 1500°C  with step 50°C. Finally, the distribution of phases
across 31 temperature points for each element composition was generated
as output.
\item \textbf{BO-driven space}: The second dataset is also of size 15,000
alloys and is created by searching via Thermo-Calc for a desired property
in FCC phase by Bayesian optimization. In alloy design, the FCC phase
proportion should be low (\textless{} 95\%) at low temperature and
high (\textgreater{} 98\%) at high temperature. We investigated the
FCC phases of known alloys at 200°C and 500°C in Fig.~\ref{fig:fcc-phases}.
It shows that these FCC phases have roughly a linear trend. Let $y_{200}$
and $y_{500}$ denote the FCC phases at 200°C and 500°C. We simply
fit a regression line $y_{500}=ay_{200}+b$ to estimate $[a,b]=[0.45,0.56]$.
\\
\\
We use Bayesian optimization and search in the domain of 9 auxiliary
elements $E=\left\{ \text{Cr, Cu, Mg, Ti, Zn, Zr, Mn, Si, Ni}\right\} $
for the phases close to this regression line, the shaded region in
Fig.~\ref{fig:fcc-phases}. The objective for the search is given
in Appendix~\ref{sec:Objective-for-BO-dataset}. The Bayesian algorithm
inputs the composition into Thermo-Calc and gets an objective as a
function of the Thermo-Calc output. We carried out the search for
one week period and yielded 120 search results. We randomly collect
1,000 trajectory points satisfying the mentioned condition and vary
their composition by $\pm20\%$, each point 15 times, resulting in
a dataset of size 15,000 data points.
\end{itemize}
For our inverse problem, we used the phase diagram as input $\yb$
and trained models to predict the element composition output $\xb$.
We use $k$-fold validation, that is, the dataset is randomly split
into 5 folds, among which 4 folds are used for training and 1 fold
is used for testing. The folds are alternated and the experiment is
repeated 5 times. Mean and standard deviation of errors are reported.

\subsection{Models setup}

We use 7 different models:
\begin{enumerate}
\item \emph{Baseline 1}: A Random Forests (RF). This model fits $30$ randomized
decision trees on random sub-samples, then uses averaging to avoid
over-fitting. The max-features parameter is set to 100.
\item \emph{Baseline 2}: A standard multi-layer perceptron (MLP) with three
hidden layers of size 500, 100, and 50 respectively. The ReLU activation
function is used for the hidden layers.
\item A mixture density network (MDN). This model has three sub-networks
for the component means, variances, mixing weights respectively to
make a mixture of Gaussian as output. These networks have the same
hidden sizes as the MLP above.
\item The proposed conditional variational autoencoder MLP (CVAE-MLP). There
are three components corresponding to the three conditional distributions
(Eq.~(\ref{eq:predict-x}): 1) the recognition model $P(\zb|\vb)$
to estimate the hidden posterior, this model has inside two sub-networks
for the mean and standard deviation of the latent vector $\zb$; 2)
the generation network $P(\hb|\vb,\zb)$ to generate the unspecified
property given the specified one and random noise; and 3) the prediction
network $P(\xb|\vb,\hb)$ to predict the design parameters. These
component networks have similar hidden sizes to the above MLP except
that the second component has reverse order of layers. The latent
size 30.
\item The proposed CVAE-MDN. This model is similar to CVAE-MLP except the
prediction network is MDN.
\item The alternative conditional generative adversarial network MLP (CGAN-MLP).
This also has three components: 1) the generator $P(\hb|\vb,\zb)$;
2) the discriminator to decide whether $\hb$ is real; and 3) the
prediction network $P(\xb|\vb,\hb)$. The networks also have similar
hidden sizes as above.
\item The CGAN-MDN. This model is similar to CGAN-MLP except the prediction
network is MDN.
\end{enumerate}
All neural network models were trained using ADAM optimizer with learning
rate 0.001 until convergence with mini-batch size of 50. 

\subsection{Performance measures \label{subsec:Performance-measures}}

Alloys vary in their composition, not just the proportion of elements,
but also the existence of elements. On average the proportion of zeros
of the alloy elements is about 20\% in our dataset. That is, each
alloy is composed of about 8 out of 10 elements. Given this fact,
we reported two prediction errors: Relative Error for those elements
which are nonzero and Absolute Error for those which are zero. Let
$\xb_{i}=\left(x_{i1},x_{i2},...,x_{iM}\right)$ be the true composition
for instance $i$, and $\hat{\xb}_{i}$ be the predicted composition.
Then the relative error (for non-zero element $x_{ij}>0$) and absolute
error (for zero element $x_{ij}=0$) are defined by:

\begin{equation}
r_{ij}=\frac{|\hat{x}_{ij}-x_{ij}|}{x_{ij}};\qquad a_{ij}=\hat{x}_{ij}\label{eq:relative-error}
\end{equation}
The relative and absolute errors for the test set are computed as:

\[
r=\frac{1}{N}\sum_{i=1}^{N}\frac{1}{M_{i1}}\sum_{j=1}^{M}r_{ij};\qquad a=\frac{1}{N}\sum_{i=1}^{N}\frac{1}{M_{i0}}\sum_{j=1}^{M}a_{ij}
\]
where $M_{i0}$, $M_{i1}$ are the number of zeros and non-zeros elements
of $\xb_{i}$ respectively.

\begin{table*}
\begin{centering}
\begin{tabular}{|c|c|c|c|c|}
\hline 
 & \multicolumn{2}{c|}{Known-alloy dataset} & \multicolumn{2}{c|}{BO-search dataset}\tabularnewline
\hline 
\textbf{Method} & \textbf{Relative (\%)} & \textbf{Absolute (\%)} & \textbf{Relative (\%)} & \textbf{Absolute (\%)}\tabularnewline
\hline 
\hline 
RF & $3.21\pm0.02$ & $0.00\pm0.00$ & $6.37\pm2.13$ & $0.01\pm0.00$\tabularnewline
\hline 
MLP & $1.10\pm0.03$ & $0.00\pm0.00$ & $3.41\pm1.48$ & $0.01\pm0.01$\tabularnewline
\hline 
MDN & $0.52\pm0.00$ & $0.00\pm0.00$ & $2.95\pm1.32$ & $0.00\pm0.01$\tabularnewline
\hline 
\end{tabular} 
\par\end{centering}
\caption{Errors for completed phases input, averaged across 5 folds. For MDN,
the mode with lowest error is reported. \label{tab:30-alloys} }
\end{table*}

\subsection{Prediction accuracy}

We carried out several experimental settings to study the model ability
in predicting alloy compositions. In the first experiment, we tried
the prediction using the full phase matrix as input, see Fig.~\ref{fig:phase-diagrams}.
In the following experiments, we compared different models and investigated
whether they can generalize given an imprecise input.

\begin{table*}
\begin{centering}
\begin{tabular}{|c|c|c|c|c|}
\hline 
 & \multicolumn{2}{c|}{Known-alloy dataset} & \multicolumn{2}{c|}{BO-search dataset}\tabularnewline
\hline 
\textbf{Method} & \textbf{Relative (\%)} & \textbf{Absolute (\%)} & \textbf{Relative (\%)} & \textbf{Absolute (\%)}\tabularnewline
\hline 
\hline 
RF & $4.38\pm0.01$ & $0.00\pm0.00$ & $8.49\pm1.34$ & $0.01\pm0.01$\tabularnewline
\hline 
MLP & $3.43\pm0.07$ & $0.00\pm0.00$ & $11.91\pm2.54$ & $0.03\pm0.02$\tabularnewline
\hline 
MDN & $2.28\pm0.22$ & $0.00\pm0.00$ & $7.83\pm1.11$ & $0.01\pm0.01$\tabularnewline
\hline 
CVAE-MLP & $2.50\pm0.24$ (a) & $0.00\pm0.00$ & $7.42\pm2.03$ (e) & $0.01\pm0.01$ (i)\tabularnewline
\hline 
CVAE-MDN & $2.08\pm0.12$ (b) & $0.00\pm0.00$ & $4.23\pm0.67$ (f) & $0.00\pm0.00$ (j)\tabularnewline
\hline 
CGAN-MLP & $3.18\pm0.18$ (c) & $0.00\pm0.00$ & $8.39\pm2.33$ (g) & $0.00\pm0.00$ (k)\tabularnewline
\hline 
CGAN-MDN & $2.30\pm0.18$ (d) & $0.00\pm0.00$ & $7.38\pm0.70$ (h) & $0.00\pm0.00$ (l) \tabularnewline
\hline 
\end{tabular} 
\par\end{centering}
\caption{Errors for partial phases input (50\% phases missing). (a) ave: $3.41\pm0.26$,
max: $4.45\pm0.28$; (b) ave: $2.50\pm0.11$, max: $3.34\pm0.39$;
(c) ave: $3.33\pm0.20$, max: $3.5\pm0.22$ (d) ave: $2.58\pm0.17$,
max: $4.52\pm0.15$; (e) ave: $8.62\pm1.83$, max: $10.71\pm2.76$;
(f) ave:$6.28\pm1.34$, max: $8.82\pm3.52$; (g) ave:$10.45\pm3.17$,
max: $12.85\pm6.46$; (h) ave: $9.68\pm0.48$, max: $11.39\pm2.22$;
(i) ave: $0.01\pm0.01$, max: $0.01\pm0.01$; (j) ave: $0.0\pm0.0$,
max: $0.01\pm0.01$; (k) ave: $0.02\pm0.01$, max: $0.05\pm0.07$;
(l) ave: $0.02\pm0.02$, max: $0.02\pm0.04$. \label{tab:partial}.}
\end{table*}

\paragraph{Predicting element compositions}

Table~\ref{tab:30-alloys} shows the error rate for this task using
RF, MLP, and MDN. With the present of full phase matrices, predicted
compositions has low error rates. For the known-alloy dataset RF can
achieve 3.21\% relative error, while MLP performs better at 1.1\%.
MDN has the lowest error, 0.5\%. For the BO-search dataset RF, MLP
and MDN achieve 6.37\%, 3.41\%, and 2.95\% relative errors respectively.
This supports the hypothesis that the output have multiple modes.

\paragraph{Predicting element compositions using partially known phases}

Table~\ref{tab:partial} compares different methods in predicting
element compositions when the input phases are partially known, only
50\% of the phases are presented to the model. Since the CVAE layer
outputs a distribution instead of a single vector, we take 20 samples
$\zb$'s ($\zb\sim\mathcal{N}(\boldsymbol{0},I)$) for each partial
input $\vb$ and output 20 fully reconstructed $\bar{\hb}$'s at the
CVAE layer. Then these 20 reconstructions, combined with $\vb$, are
then passed to the MLP/MDN layer to predict 20 output $\bar{\xb}$'s.
The minimum, mean, and maximum errors for these $\bar{\xb}$'s against
the single target $\xb$ are reported. The effect of reconstruction
using CVAE is clearly demonstrated. For the first dataset the error
of MLP drops by 0.93\%, from $3.43\%$ (without reconstruction) to
$2.50\%$ (with reconstruction). Likewise, the error of MDN drops
from $2.28\%$ (without reconstruction) to $2.08\%$ (with reconstruction).
Similarly, for the second dataset the error of MLP drops by 4.49\%,
from 11.91\% to 7.42\%. The error of MDN drops by 3.6\%, from 7.83\%
to 4.23\%. The CGAN-based models also reduce the errors for the basic
MLP and MDN but not as good as the CVAE-based models.

We also experimented with different missing phase ratio for the MLP
model and found that the error rate is still low (\textless 10\%)
even at 70\% missing rate. Only at 80\% missing rate the error becomes
significant (about 70\%). This suggests that there is a small number
of phases (mainly the compounds) that contains most of the element
composition information.

\paragraph{Example of phase diagrams having mixture outputs}

Given an imprecise input (a partial designed phase diagram) there
can be multiple alloy compositions satisfying this design. Fig.~\ref{fig:fcc-phases}
depicts some examples where a design FCC phase at 200°C and 500°C
has several possible alloy compositions. The data points are very
close in this FCC plots. For an example, let the design phase be ($\text{fcc}_{200}=0.915$,
$\text{fcc}_{500}=0.99$), then this is satisfied by three different
alloys 7075, 7076, and 7175. The 7075 and 7175 have Cr proportion
of 0.55 and no Mn element. while the 7076 has Mn=0.33 and no Cr. Also
the 7075 has double the Ti proportion compared to the 7175. Similarly,
for the design phase with ($\text{fcc}_{200}=0.985$, $\text{fcc}_{500}=0.998$)
there are several 6000 series alloys satisfying it.

\subsection{Verifying the error in Thermo-Calc}

Although the prediction of design parameters have been fairly accurate
with $2-3\%$ relative error, it is possible that the design may be
at the edge of the plausible design region. Thus it is important to
verify whether we can reconstruct the original specification given
the prediction using the simulator itself. In this experiment, the
predicted element compositions (20 output samples for one partial
input phase diagram) for the partially known phases are fed back into
Thermo-Calc for simulating the new phase diagrams. The phase diagrams,
combined from the predicted property $\bar{\hb}$ and the specified
property $\vb$, are then compared to the original phase diagram $\yb$
to check whether the variational method has produced consistent elements
for these phases. We expect the observed error to be small.

The absolute error for 1 phase diagram and 1 Thermo-Calc simulated
diagram from the predicted alloy (using MLP) is: $\left\Vert \hb-\bar{\hb}\right\Vert =0.03$
(each phase of the diagram is scaled to have the global range $[0,1]$
across the dataset). The relative and absolute errors for observed
phases between one partial input phase diagram and 20 simulated diagram
from 20 predicted alloy compositions are 2.94\% and 0.00\% respectively
(3,000 test examples). The relative error for phases are calculated
similarly to the relative error for element compositions for nonzero
phases in Section~\ref{subsec:Performance-measures}.

We rerun phase simulations in the Thermo-Calc software given the CVAE-MDN
predicted composition outputs, then observe the error of reconstruction.
The minimum errors (\%) of 14 observed phases are shown in Table~\ref{tab:cvae-mdn-per-phase-error}.
\begin{center}
\begin{table}
\begin{centering}
\begin{tabular}{|l|c|c|}
\hline 
\textbf{Phase} & \textbf{Relative} & \textbf{Absolute}\tabularnewline
\hline 
\hline 
AL12MN & 0.52  & 0.0\tabularnewline
\hline 
AL13CR4SI4 & 0.95  & 0.0\tabularnewline
\hline 
AL15SI2M4 & 6.42  & 0.0\tabularnewline
\hline 
AL28CU4MN7 & 9.41  & 0.0\tabularnewline
\hline 
AL2CU\_C16 & 3.63  & 0.0\tabularnewline
\hline 
AL31MN6NI2 & 2.04  & 0.0\tabularnewline
\hline 
AL3NI2 & 0.64  & 0.0\tabularnewline
\hline 
AL3NI\_D011 & 0.95  & 0.0\tabularnewline
\hline 
AL3TI\_D022 & 2.15  & 0.0\tabularnewline
\hline 
AL3ZR\_D023 & 1.72  & 0.0\tabularnewline
\hline 
AL45V7 & 1.46  & 0.0\tabularnewline
\hline 
AL6MN & 2.94  & 0.0\tabularnewline
\hline 
AL7CU4NI & 2.89  & 0.0\tabularnewline
\hline 
ALMG\_BETA & 5.51  & 0.0\tabularnewline
\hline 
\hline 
\textbf{Average} & 2.94 & 0.0\tabularnewline
\hline 
\end{tabular} 
\par\end{centering}
\caption{Minimum Errors (\%) of 14 observed phases when rerunning phase simulations
in the Thermo-Calc software, from the CVAE-MDN predicted composition
outputs. \label{tab:cvae-mdn-per-phase-error}}
\end{table}
\par\end{center}

\subsection{Computational efficiency and comparisons with the search-based methods}

Recall that we can, in theory, perform extensive search for the best
design parameters $\xb$ for a given target properties $\yb$ by running
repeatedly the simulator to produce $\yb$ from each $\xb$. The time
cost is the product of the time per simulator call, and the total
number of calls. For this we compare our method with popular black-box
search methods (i.e., genetic algorithm) for this inverse design problem.
As a demonstration, we carried out the search for a random desired
phase diagram, its true composition is \{Cr=0.24, Mg=2.96, Ti=0.06,
Zn=4.29, Mn=0.2, Si=0.12, Al=92.14\}. For this, the trained CVAE-MDN
gives immediately (in milliseconds) a set of 20 compositions, and
accordingly 20 simulated phase diagrams by Thermo-Calc with a minimum
of 0.7\%, a mean of 2.0\% , and a maximum of 3.36\% relative error
respectively.

\begin{figure}
\begin{centering}
\includegraphics[width=1\columnwidth]{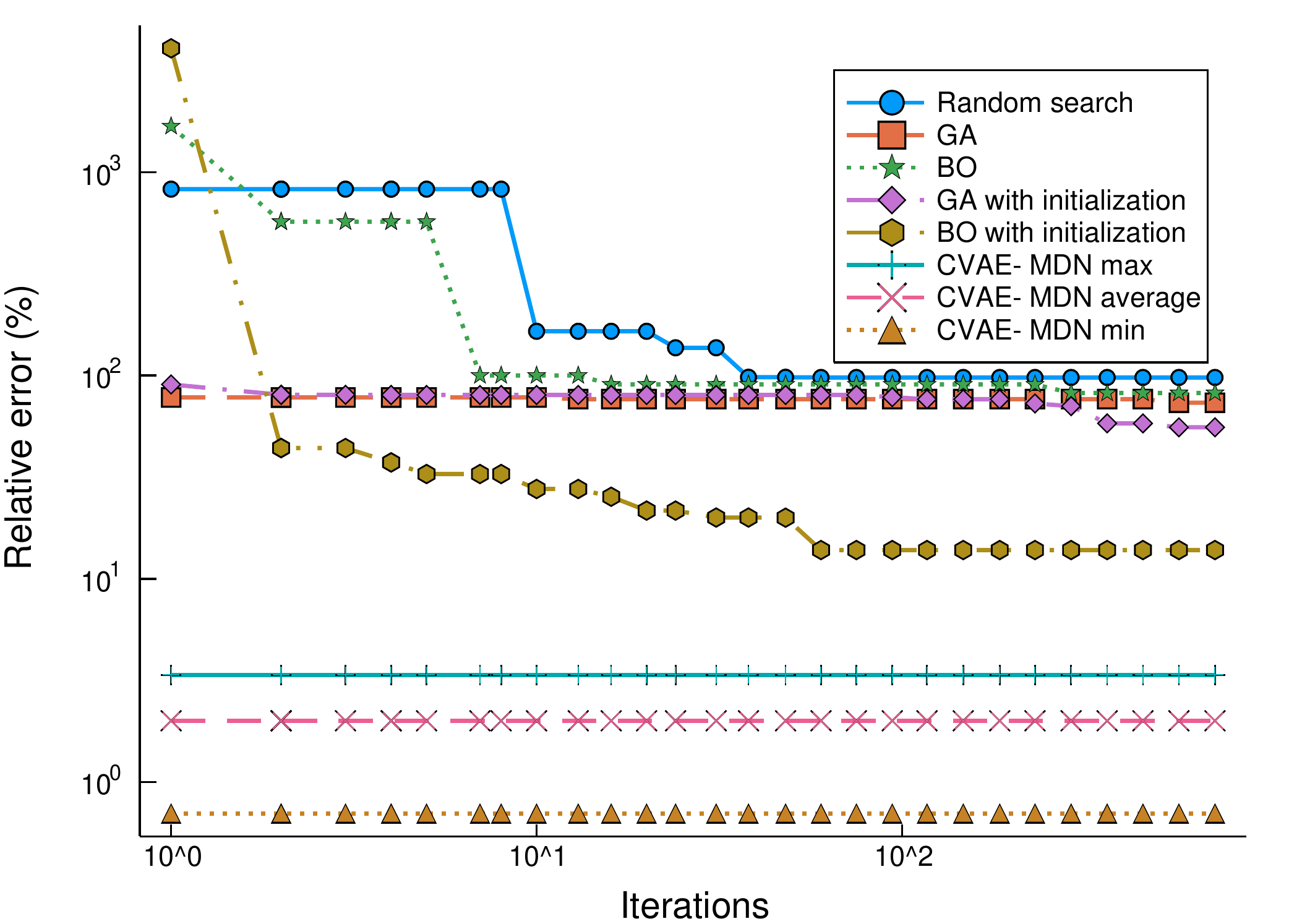}
\par\end{centering}
\caption{Error versus search time comparisons. The search did not reach any
meaningful solution after 1000 (expensive) steps. This is opposed
to meaningful machine learning prediction in one (cheap) step. \label{fig:Error-vs-search-time}}
\end{figure}

Fig.~\ref{fig:Error-vs-search-time} shows the error versus search
time for the random search, genetic algorithm (GA), and Bayesian optimization
(BO). Random search fails to get any noticeable improvement after
1000 iterations. GA shows progress but after 1000 steps, the relative
error is still far above the 50\% mark. BO shows better performance
than random search and GA.

It took CVAE-MDN 1.825ms to find a solution without any search. Thermo-Calc
took 80s for one run but the simulation-based search did not find
any significant solution after 1000 runs (10+ hours, which is already
3-order of magnitude slower). We expect that it may takes days to
achieve solutions of similar quality that is found with our learning-based
solutions. The speed up by learning-based techniques (our methods)
can be increased many folds \emph{for free} since our methods can
be run in batch (e.g., 100 examples) without a significant increase
in time, thanks to the modern GPU architectures. For example, a batch
of 30 alloys took only 2ms on the GPU in our experiment. The experiments
were taken on the system with Intel Xeon E5-1650 v4 3.60 GHz CPU,
Nvidia GeForce GTX 1080 Ti GPU, and 32GB RAM.

\subsection{Alloy distribution and generation}

\begin{figure*}
\begin{centering}
\includegraphics[width=1\textwidth]{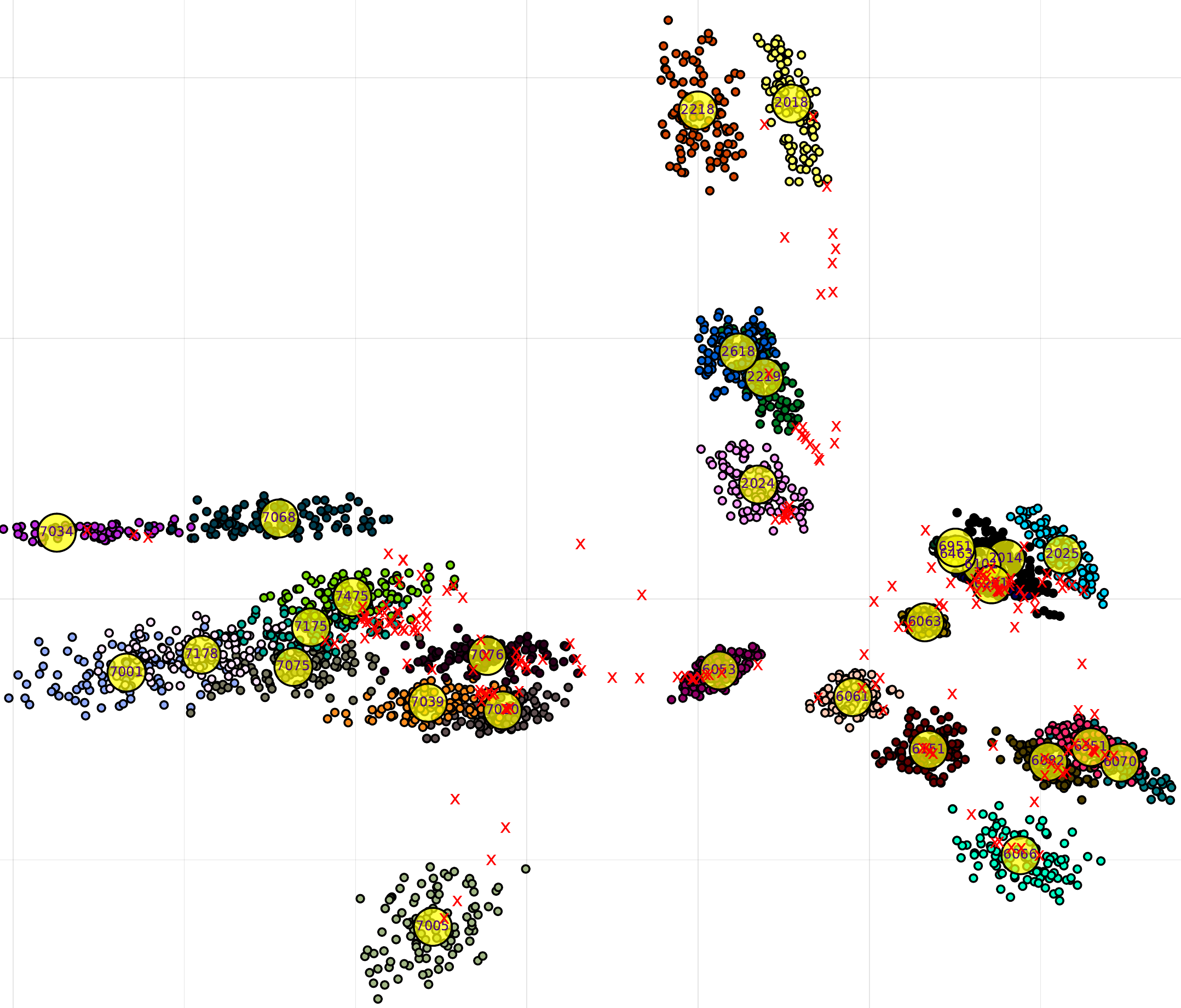}
\par\end{centering}
\caption{Alloy distribution using PCA. Each point represents an alloy varying
around a known alloy (yellow circle); the color represents the alloy
class. The red crosses are some examples generated from the CVAE-MDN
model at test time, showing the model can generate both compositions
that are near the known alloys and unexplored ones that are interpolations
of those in the training data. Best viewed in color. \label{fig:alloy-distribution}}
\end{figure*}

Fig.~\ref{fig:alloy-distribution} visualizes the data distribution.
It also shows some examples generated by the CVAE-MDN method. The
generated compositions are either near the known alloys or in the
unexplored region between those alloys in the training data.

\section{Conclusion}

We have proposed a novel data--driven framework for designing new
materials and products. Using the power of data mining, our framework
has shifted the current design paradigm of searching for the target
design to instant prediction. In day-to-day life of an experimental
designer, this may easily bring a speed up of thousands (if not millions)
of times. Although in this paper, we have demonstrated the applicability
of our work mainly for designing alloys, our work is broadly applicable
to most of the scientific design applications where it is possible
to query simulators in offline mode or collect experimental design
data in an ongoing basis. In the future, it would be useful to advance
the scope of this research further by allowing qualitative specification
of target properties (e.g. allowing specifications in term of preferences
of component). Additionally, since the basic physical and chemical
laws are often shared across different class of products, it may be
interesting to use transfer learning across different product classes
to achieve better prediction accuracy with smaller sized datasets.
Finally, as a data-driven approach, generating new samples outside
the training space is difficult, suggesting a new way to capture higher-order,
compositional structures with proper uncertainty quantification.

\bibliographystyle{plain}

\begin{thebibliography}{10}

\bibitem{agrawal2016perspective}
Ankit Agrawal and Alok Choudhary.
\newblock Materials informatics and big data: Realization of the "fourth
  paradigm" of science in materials science.
\newblock {\em Apl Materials}, 4(5):053208, 2016.

\bibitem{aspuru2018materials}
Al{\'a}n Aspuru-Guzik and Kristin Persson.
\newblock Materials acceleration platform: Accelerating advanced energy
  materials discovery by integrating high-throughput methods and artificial
  intelligence.
\newblock 2018.

\bibitem{bishop1994mixture}
Christopher Bishop.
\newblock Mixture density networks.
\newblock {\em Technical Report}, 1994.

\bibitem{gani2002property}
Rafiqul Gani and Efstratios~N Pistikopoulos.
\newblock Property modelling and simulation for product and process design.
\newblock {\em Fluid Phase Equilibria}, 194:43--59, 2002.

\bibitem{gomez2018automatic}
Rafael G{\'o}mez-Bombarelli, Jennifer~N Wei, David Duvenaud, Jos{\'e}~Miguel
  Hern{\'a}ndez-Lobato, Benjam{\'\i}n S{\'a}nchez-Lengeling, Dennis Sheberla,
  Jorge Aguilera-Iparraguirre, Timothy~D Hirzel, Ryan~P Adams, and Al{\'a}n
  Aspuru-Guzik.
\newblock Automatic chemical design using a data-driven continuous
  representation of molecules.
\newblock {\em ACS central science}, 4(2):268--276, 2018.

\bibitem{goodfellow2014generative}
Ian Goodfellow, Jean Pouget-Abadie, Mehdi Mirza, Bing Xu, David Warde-Farley,
  Sherjil Ozair, Aaron Courville, and Yoshua Bengio.
\newblock Generative adversarial nets.
\newblock In {\em Advances in Neural Information Processing Systems}, pages
  2672--2680, 2014.

\bibitem{gupta2015structure}
Akash Gupta, Ahmet Cecen, Sharad Goyal, Amarendra~K Singh, and Surya~R
  Kalidindi.
\newblock Structure--property linkages using a data science approach:
  application to a non-metallic inclusion/steel composite system.
\newblock {\em Acta Materialia}, 91:239--254, 2015.

\bibitem{hey2009fourth}
Tony Hey, Stewart Tansley, Kristin~M Tolle, et~al.
\newblock {\em The fourth paradigm: data-intensive scientific discovery},
  volume~1.
\newblock Microsoft research Redmond, WA, 2009.

\bibitem{kalidindi2015materials}
Surya~R Kalidindi and Marc De~Graef.
\newblock Materials data science: current status and future outlook.
\newblock {\em Annual Review of Materials Research}, 45:171--193, 2015.

\bibitem{kaufman1970computer}
Larry Kaufman and Harold Bernstein.
\newblock Computer calculation of phase diagrams. with special reference to
  refractory metals.
\newblock 1970.

\bibitem{kingma2013auto}
Diederik~P Kingma and Max Welling.
\newblock {Auto-encoding variational Bayes}.
\newblock {\em arXiv preprint arXiv:1312.6114}, 2013.

\bibitem{le2018variational}
Hung Le, Truyen Tran, Thin Nguyen, and Svetha Venkatesh.
\newblock Variational memory encoder-decoder.
\newblock {\em NIPS}, 2018.

\bibitem{li2012using}
XM~Li and JJ~Yu.
\newblock Using orthogonal experimental design to optimize alloy composition.
\newblock In {\em IOP Conference Series: Materials Science and Engineering},
  volume~33, page 012065. IOP Publishing, 2012.

\bibitem{limbourg2005optimization}
Philipp Limbourg and Daniel E~Salazar Aponte.
\newblock An optimization algorithm for imprecise multi-objective problem
  functions.
\newblock In {\em Evolutionary Computation, 2005. The 2005 IEEE Congress on},
  volume~1, pages 459--466. IEEE, 2005.

\bibitem{liu2017materials}
Yue Liu, Tianlu Zhao, Wangwei Ju, and Siqi Shi.
\newblock Materials discovery and design using machine learning.
\newblock {\em Journal of Materiomics}, 3(3):159--177, 2017.

\bibitem{lukas2007computational}
Hans~Leo Lukas, Suzana~G Fries, Bo~Sundman, et~al.
\newblock {\em Computational thermodynamics: the Calphad method}, volume 131.
\newblock Cambridge university press Cambridge, 2007.

\bibitem{maddison2014sampling}
Chris~J Maddison, Daniel Tarlow, and Tom Minka.
\newblock A* sampling.
\newblock In {\em Advances in Neural Information Processing Systems}, pages
  3086--3094, 2014.

\bibitem{mattei2018leveraging}
Pierre-Alexandre Mattei and Jes Frellsen.
\newblock Leveraging the exact likelihood of deep latent variables models.
\newblock {\em arXiv preprint arXiv:1802.04826}, 2018.

\bibitem{mirza2014conditional}
Mehdi Mirza and Simon Osindero.
\newblock Conditional generative adversarial nets.
\newblock {\em arXiv preprint arXiv:1411.1784}, 2014.

\bibitem{nazabal2018handling}
Alfredo Nazabal, Pablo~M Olmos, Zoubin Ghahramani, and Isabel Valera.
\newblock {Handling incomplete heterogeneous data using VAEs}.
\newblock {\em arXiv preprint arXiv:1807.03653}, 2018.

\bibitem{norskov2011density}
Jens~K N{\o}rskov, Frank Abild-Pedersen, Felix Studt, and Thomas Bligaard.
\newblock Density functional theory in surface chemistry and catalysis.
\newblock {\em Proceedings of the National Academy of Sciences},
  108(3):937--943, 2011.

\bibitem{pham2018graph}
Trang Pham, Truyen Tran, and Svetha Venkatesh.
\newblock Graph memory networks for molecular activity prediction.
\newblock {\em ICPR}, 2018.

\bibitem{qian1993process}
Yu~Qian, Patrick Tessier, and Guy~A Dumont.
\newblock Process modelling and optimization of systems with imprecise and
  conflicting equations.
\newblock {\em Engineering Applications of Artificial Intelligence},
  6(1):39--47, 1993.

\bibitem{raccuglia2016machine}
Paul Raccuglia, Katherine~C Elbert, Philip~DF Adler, Casey Falk, Malia~B Wenny,
  Aurelio Mollo, Matthias Zeller, Sorelle~A Friedler, Joshua Schrier, and
  Alexander~J Norquist.
\newblock Machine-learning-assisted materials discovery using failed
  experiments.
\newblock {\em Nature}, 533(7601):73, 2016.

\bibitem{ramprasad2017machine}
Rampi Ramprasad, Rohit Batra, Ghanshyam Pilania, Arun Mannodi-Kanakkithodi, and
  Chiho Kim.
\newblock Machine learning in materials informatics: recent applications and
  prospects.
\newblock {\em npj Computational Materials}, 3(1):54, 2017.

\bibitem{rezende2015variational}
Danilo~Jimenez Rezende and Shakir Mohamed.
\newblock Variational inference with normalizing flows.
\newblock {\em arXiv preprint arXiv:1505.05770}, 2015.

\bibitem{rezende2014stochastic}
Danilo~Jimenez Rezende, Shakir Mohamed, and Daan Wierstra.
\newblock Stochastic backpropagation and variational inference in deep latent
  gaussian models.
\newblock In {\em International Conference on Machine Learning}, volume~2,
  2014.

\bibitem{sanchez2018inverse}
Benjamin Sanchez-Lengeling and Al{\'a}n Aspuru-Guzik.
\newblock {Inverse molecular design using machine learning: Generative models
  for matter engineering}.
\newblock {\em Science}, 361(6400):360--365, 2018.

\bibitem{schafer1999multiple}
Joseph~L Schafer.
\newblock Multiple imputation: a primer.
\newblock {\em Statistical methods in medical research}, 8(1):3--15, 1999.

\bibitem{shahriari2016taking}
Bobak Shahriari, Kevin Swersky, Ziyu Wang, Ryan~P Adams, and Nando De~Freitas.
\newblock {Taking the human out of the loop: A review of Bayesian
  optimization}.
\newblock {\em Proceedings of the IEEE}, 104(1):148--175, 2016.

\bibitem{shu2017bottleneck}
Rui Shu, Hung~H Bui, and Mohammad Ghavamzadeh.
\newblock Bottleneck conditional density estimation.
\newblock In {\em International Conference on Machine Learning}, pages
  3164--3172, 2017.

\bibitem{sohn2015learning}
Kihyuk Sohn, Honglak Lee, and Xinchen Yan.
\newblock Learning structured output representation using deep conditional
  generative models.
\newblock In {\em Advances in Neural Information Processing Systems}, pages
  3483--3491, 2015.

\bibitem{trippe2018conditional}
Brian~L Trippe and Richard~E Turner.
\newblock Conditional density estimation with bayesian normalising flows.
\newblock {\em arXiv preprint arXiv:1802.04908}, 2018.

\bibitem{weissman2014design}
Steven~A Weissman and Neal~G Anderson.
\newblock Design of experiments (doe) and process optimization. a review of
  recent publications.
\newblock {\em Organic Process Research \& Development}, 19(11):1605--1633,
  2014.

\bibitem{yoon2018gain}
Jinsung Yoon, James Jordon, and Mihaela van~der Schaar.
\newblock {GAIN: Missing Data Imputation using Generative Adversarial Nets}.
\newblock {\em arXiv preprint arXiv:1806.02920}, 2018.

\bibitem{zunger2018inverse}
Alex Zunger.
\newblock Inverse design in search of materials with target functionalities.
\newblock {\em Nature Reviews Chemistry}, 2:0121, 2018.

\end{thebibliography}

\appendix

\section{Appendix \label{sec:Objective-for-BO-dataset}}

Let $\xb$ denote the alloy composition and $\yb$ the phases. The
Bayesian Optimisation searches in the space of alloy ($\xb$) to minimize
objective defined in the phase space ($\yb$). The compositions are
constrained by:
\[
\sum_{i\in E}x_{i}\le15\%
\]
The objective is

\[
\text{min}_{\yb}L(\yb)=L_{1}(\yb)+L_{2}(\yb)+0.1L_{3}(\yb)-0.01L_{4}(y\yb)
\]
where
\begin{itemize}
\item $L_{1}(\yb)$ is defined as: 
\[
L_{1}(\yb)=\begin{cases}
0 & d_{1}<0.05\\
d_{1}^{2} & \text{otherwise}
\end{cases}
\]
where $d_{1}$ is the distance to the regression line $f_{\theta}$
\item $L_{2}(\yb)$ encourages search in the region right to the cut off
barriers at 200°C, which is defined as:
\end{itemize}
\[
L_{2}(\yb)=\begin{cases}
0 & d_{2}(\yb)>0\\
d_{2}^{2} & \text{otherwise}
\end{cases}
\]

where $d_{2}=y_{200}-0.88$ is the distance to the line $y_{200}=0.88$
\begin{itemize}
\item $L_{3}(\yb)$ encourages the diversity of the search:
\[
L_{3}(\yb)=\text{max}_{\yb'\in\text{S}}\text{exp}\left(-\frac{1}{\sigma^{2}}\left\Vert \yb'-\yb\right\Vert ^{2}\right)
\]
where $S=\{\yb\}$ the set of all BO visited points.
\item $L_{4}(\yb)$ encourages nonzero elements:
\[
L_{4}(\yb)=\left|\left\{ y_{i}:y_{i}>10^{-6},i\in E\right\} \right|.
\]
\end{itemize}

\end{document}